\documentclass[graybox]{svmult}
\usepackage{svg}

\usepackage[justification=centering]{caption}
\usepackage{mathptmx}       
\usepackage{helvet}         
\usepackage{courier}        
\usepackage{type1cm}        
\usepackage{makeidx}         
\usepackage{graphicx}        
\usepackage{multicol}        
\usepackage[bottom]{footmisc}

\makeindex             

\begin{document}

\title*{Deep End-to-end Fingerprint Denoising and Inpainting}
\author{Youness MANSAR}
\institute{Fortia Financial Solutions \at 17 avenue George V, Paris, \email{youness.mansar@fortia.fr}}
%
%
\maketitle

\abstract{This work describes our winning solution for the Chalearn LAP In-painting Competition Track 3 - Fingerprint Denoising and In-painting. The objective of this competition is to reduce noise, remove the background pattern and replace missing parts of fingerprint images in order to simplify the verification made by humans or third-party software.
In this paper, we use a U-Net like CNN model that performs all those steps end-to-end after being trained on the competition data in a fully supervised way.
This architecture and training procedure achieved the best results on all three metrics of the competition \footnote{\url{http://chalearnlap.cvc.uab.es/challenge/26/track/32/result/}}.}

\section{Background}
Fingerprints play an important role in privacy and identity verification but can also be used in forensic operations. This means that having the ability to accurately process and match fingerprints can be a valuable asset. This is what motivates this work where the objective is to retrieve a cleaned image of a fingerprint out of a noisy, distorted version.\newline
Generally, images contain noise and perturbations that may be due to the acquisition device, compression method or post-processing done. This motivates the research on tools and methods like denoising and inpainting to alleviate this problem. They are used as a pre-processing step in order to simplify the subsequent tasks and improve the target performance. In our case, the end goal is to improve the fingerprint false acceptance rate or the false rejection rate.\newline
One approach to denoising is the TV method  \cite{Osher05aniterative} which is based on the principle that noisy images have a high total variation, the aim of the TV approach is to thus reduce the regularized total variation of the input image.\newline
\cite{Buades05areview} reviews multiples methods to denoising like the Gaussian smoothing model or translation invariant wavelet thresholding, among others.\newline
A more recent direction to denoising and inpainting is based on deep neural networks where a sequence of convolution layers are optimized to learn a mapping from a noisy image to a "clean" version of that image. \cite{DBLP:journals/corr/SvobodaMB17} studies the same problem as the Chalearn competition and uses and proposes encoder-decoder architecture to solve it. \cite{DBLP:journals/corr/MaoSY16a} shows that using skip connections helps avoid the issues related to training deep neural networks like the vanishing gradient problem.\newline
Similar to \cite{DBLP:journals/corr/MaoSY16a}, \cite{DBLP:journals/corr/RonnebergerFB15} introduces an architecture called U-net that is also an encoder-decoder type with skip connections that is used primarily for image segmentation. U-Net showed impressive results when used along with data augmentation even when the size of the dataset is small. In this work, we are going to use an architecture that is similar to U-Net and show that it can be applied successfully even outside pure segmentation tasks.
\section{Data}
The dataset provided by the organizers consisted of 84000 (200, 400) fingerprint images generated using Anguli: Synthetic Fingerprint Generator. Those images were then artificially degraded by adding a background and random transformations (blur, brightness, contrast, elastic transformation, occlusion, scratch, resolution, rotation).
The objective is to retrieve the clean fingerprint image from the degraded version. We use the set of parallel data (Degraded image, Clean image) as the (Input, Ground Truth) of our model training.
\section{Proposed solution}
\subsection{Model}
\begin{figure}[h]
\includegraphics[width=0.65\textwidth, height=0.35\textwidth]{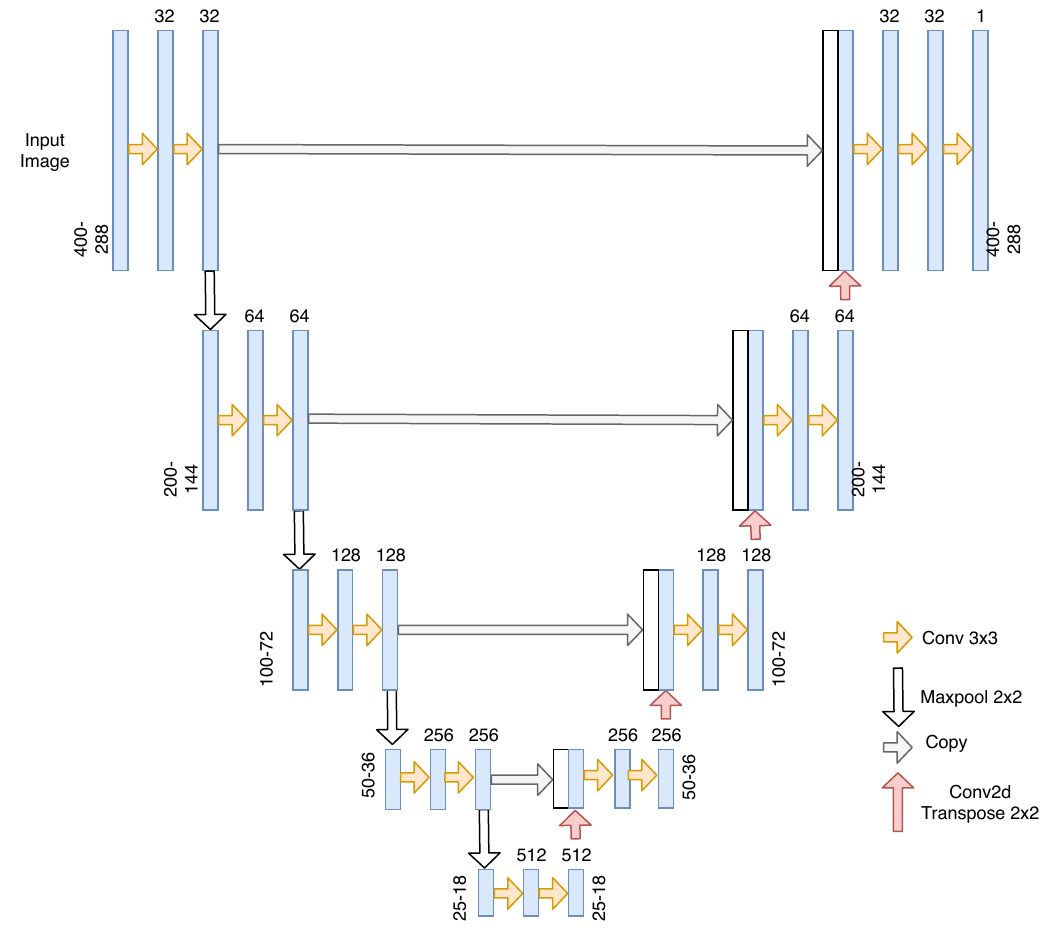}
\centering
\captionsetup{justification=centering}
\caption{U-net architecture used}
\label{fig:unet}
\end{figure}
The architecture used is described in Figure \ref{fig:unet} and is similar to the one introduced in \cite{DBLP:journals/corr/MaoSY16a}, except that we pad the input with zeros instead of mirroring the edges. The major advantage of this architecture is its ability to take into account a wider context when making a prediction for a pixel. This is thanks to the large number of channels used in the up-sampling operation.

\subsection{Image processing}
\textbf{Input image processing} : We apply this sequence of processing steps before feeding it to the CNN.
\begin{itemize}
    \item Normalization : we divide pixel intensities by 255 so they are in the 0-1 range.
    \item Re-sizing : The network expects each dimension of the input image to be divisible by $2^4$ because of the pooling operations.
    \item Data augmentation : Random flip (Horizontal or vertical or both), Random Shear, Random translation (Horizontal or vertical or both), Random Zoom, Random Contrast change, Random Saturation change, Random Rotation. Performed during training only.
\end{itemize}
\textbf{Output image processing} : We apply this sequence of processing steps before submitting the results.
\begin{itemize}
    \item Min-Max scaling : We min-max scale the output to the 0-255 range.
    \item Re-sizing : We re-size the size of the output to the original size of the input.
\end{itemize}

\subsection{Training Procedure}
We use Adam \cite{DBLP:journals/corr/KingmaB14} optimizer with an initial learning rate of $1e^{-4}$ that is reduced by a factor of 0.5 each time the validation loss plateaued for more than 3 epochs and the learning is stopped if the validation loss does not improve for the last 5 epochs.\newline
Implementation was done using Keras \cite{Chollet2015keras} with Tensorflow \cite{DBLP:journals/corr/AbadiABBCCCDDDG16} backend on a 1070 GTX card.
\section{Ablation study}
We do an Ablation study to understand which part of the approach helped the most.
We train on  on 7\% of available data to reduce computational costs. We report the results on the validation set
\begin{table}[h]
\centering
\begin{tabular}{p{3cm}p{1.2cm}p{1.2cm}p{1.2cm}}

Team & MAE $\downarrow$ & PSNR $\uparrow$& SSIM $\uparrow$\\
Submitted model & 0.0237 & 16.76 & 0.8085 \\ 
-Data augmentation & 0.0239 & 16.74 & 0.8085 \\ 
-Reducing lr & 0.0243 & 16.63 & 0.8010 \\
-skip connections & 0.7282 & 1.378 & 0.0001 \\ 
+pretrained vgg encoder & 0.0225 & 16.97 & 0.8142
\end{tabular}
\caption{Ablation results}
\end{table}
What helped most is reducing the learning rate while data augmentation did not help that much.
\section{Results}

\begin{table}[h]
\centering

\begin{tabular}{p{3cm}p{1.2cm}p{1.2cm}p{1.2cm}}

Team & MAE $\downarrow$ & PSNR $\uparrow$& SSIM $\uparrow$\\
CVxTz (This work) & \textbf{0.0189} & \textbf{17.6968} & \textbf{0.8427} \\
rgsl888 & 0.0231 & 16.9688 & 0.8093 \\ 
hcilab & 0.0238 & 16.6465 & 0.8033 \\ 
sukeshadigav & 0.0268 & 16.5534 & 0.8261 
\end{tabular}
\caption{Best Test results in bold}
\end{table}
Our approach gets the best results on all three metrics. Even though we only used the MAE in our loss function, it seems to have acted a good proxy for the other two metrics.\newline
As a comparison rgsl888 used a similar architecture to ours but added dilated convolutions to expand the receptive field of the network. hcilab used a hierarchical approach and sukeshadigav used an M-Net Based Convolutional Neural Network.
\section{Advantages and Limitations}
Our approach has the merit of being end-to-end, requires minimal pre-processing to the input and uses a single model. All of this simplifies the use of the approach in a real-life scenario.\newline
This approach also comes with few limitations like the fact that the train and test sets are both synthetic, which means that we do not know if the same performance will be preserved if the trained model is applied to real data. Another issue is that since the model is trained in a fully supervised way, then it is unlikely to generalize beyond the perturbations that it was trained on. This reaffirms the need to train on real fingerprint data.
\section{Conclusion and future work}
In this paper, we describe the approach we used to achieve 1st place on the Chalearn LAP In-painting Competition Track 3 - Fingerprint Denoising and In-painting. We describe the pre-processing steps needed, data augmentation, training procedure and network architecture used. We also make the code needed to reproduce the results available on github\footnote{\url{https://github.com/CVxTz/fingerprint\_denoising}}.\newline
In our future work, we will experiment with transferring representations from higher level supervised tasks or by using a semi-supervised approach like adding an adversarial loss.
%
%
%

\end{document}